\documentclass[11pt]{article}

\usepackage[preprint]{acl}

\usepackage{microtype}
\usepackage{xcolor}
\usepackage{tcolorbox}
\usepackage{multirow}
\usepackage{tabularx}
\usepackage{amsmath}
\usepackage{booktabs}
\usepackage{pdfpages}
\usepackage{colortbl}

\title{A Rising Tide Lifts All Boats: MTQE Rewards for Idioms Improve General Translation Quality}

\author{Ishika Agarwal*$^1$, Zhenlin He*$^1$, Dhruva Patil$^2$, Dilek Hakkani-Tür$^1$ \\
  $^1$UIUC, $^2$Independent  \\
  \texttt{{ishikaa2, zhenlin5, dilek}@illinois.edu, dhruvakpatil@gmail.com}
  }

\definecolor{customblue}{HTML}{0070C0}
\definecolor{customred}{HTML}{D60000}
\definecolor{custompurple}{HTML}{AC00AC}
\definecolor{custombrown}{HTML}{996600}

\begin{document}
\maketitle

\def\thefootnote{*}\footnotetext{These authors contributed equally to this work. Correspondence to \texttt{ishikaa2@illinois.edu}.}\def\thefootnote{\arabic{footnote}}

\begin{abstract}
    Non-compositional expressions (e.g., idioms, proverbs, and metaphors) pose significant challenges for neural machine translation systems because their meanings cannot be derived from individual words alone. These expressions encode rich, cultural meaning, and have both figurative and literal meanings, making accurate translation difficult. Because models are fairly good at translating compositional text, we investigate GRPO-style fine-tuning using Machine Translation Quality Estimation (MTQE) models as reward functions to train models to better translate idioms. Using Chinese and Hindi idiom datasets, we find that idiom translation abilities improve by $\sim$14 points\footnote{This is an average improvement across all metrics measuring n-gram similarity and semantic similarity. More details are available in Section \ref{sec: experiments}.}, general, non-idiomatic translation implicitly improves by $\sim$8 points, and cross-lingual translation abilities (trained on one language, evaluated on another) improves by $\sim$6 points. Overall, our work quantifies the non-compositional translation gap and offers insights for developing LLMs with stronger cross-cultural and figurative language understanding.
\end{abstract}
\maketitle

\section{Introduction}
The focus on multilingual language modeling has grown significantly \citep{adelnia2011translation,cheng2024no,finegrained2025mt,wu2018rlmt} because of how accessible LLMs become when they can converse in different languages. However, multilingual language modeling is difficult, due to alignment based problems. Broadly, these problems can be categorized into \textit{language specific knowledge} where models contain different facts about the same topic in different languages \citep{lsk, jin2025languagemodelalignmentmultilingual}, \textit{non-isomorphic} phrases or words that do not have direct translations in other languages (like the word "jugaad" in Hindi) \citep{wu2024representational, meng2025resolving}, and\textit{ non-compositional} phrases whose meaning cannot be derived from individual words) \citep{clcl}. The focus of this work is on improving the translation abilities of non-compositional sentences.

\begin{figure}
    \centering
    \includegraphics[width=\linewidth]{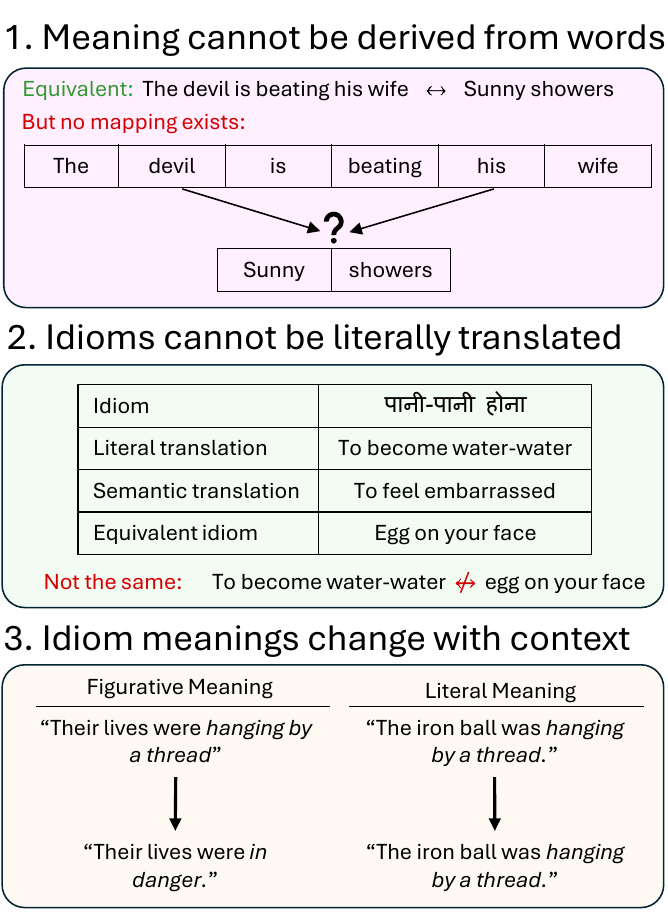}
    \caption{There are three challenges when modeling and translating non-compositional phrases.}
    \label{fig: noncomp_challenges}
\end{figure}

Specifically, translating non-compositional phrases poses three challenges, as illustrated in Figure \ref{fig: noncomp_challenges}. \textbf{Firstly}, their meanings cannot be derived from the individual, constituent words \cite{zhou2023non}. Moreover, they compress historical stories and cultural assumptions, which make their literal translations and intended meanings differ greatly.

\textbf{Second}, many idioms do not have semantically equivalent translations: language models tend to paraphrase text during translation, and thus show a strong literal translation bias towards non-compositional phrases \cite{adelnia2011translation}. These literal decodings erase the idiom's figurative intent and leads to incorrect, misleading translations.

\textbf{Third}, and finally, non-compositional phrases are highly context dependent and can have different meanings \cite{fornaciari2024hardnutcrackidiom}. These issues result in subpar translation quality, and highlight a broader cultural translation gap that current LLMs are not equipped to bridge without specialized training.

In this work, we propose to improve the translation quality of language models in the form of a training-free solution, and training-based solution. The training-free solution is a simple, three-step prompting pipeline that encourages the model to think about the cultural context and semantic meaning of an idiom before translating it.

The training-based solution uses GRPO-style fine-tuning to encourage models to translate idioms semantically. In particular, we show the efficacy of using MTQE (Machine Translation Quality Estimation) models, such as COMET \citep{rei2020comet, comet22, comet23}, as a reward model for fine-tuning LLMS with GRPO (Group Relative Policy Optimization) \cite{grpo}. MTQE models are trained on human preference data, which contains implicit signals of non-compositional phrases. We use MTQE rewards as a form of distillation to train LLMs for better translation quality. All methods are outlined in Section \ref{sec: methodology}. 

We use Chinese and Hindi idioms from existing datasets (outlined in Section \ref{sec: dataset}) to evaluate our methods. We evaluate on small language models (Qwen's 3B model and Llama's 8B model) to show that we can improve small language models for cheap and accessible translation. We measure translation quality across various dimensions of semantic and n-gram based similarity, and find that by using MTQE rewards during GRPO fine-tuning on idiomatic data:
\begin{enumerate}
    \item Idiomatic translation quality improves by $\sim$14 points (Figures \ref{fig: chinese_eval} and \ref{fig: hindi_eval}),
    \item Non-idiomatic, general, translation quality improves by $\sim$8 points (Figures \ref{fig: opus_chinese_eval} and \ref{fig: opus_hindi_eval}), and
    \item Cross-lingual semantic representation (idiom training in one language transfers to another) improves by $\sim$6 points (Figures \ref{fig: transfer_eval_hi2zh} and \ref{fig: transfer_eval_zh2hi}).
\end{enumerate}

\section{Background}
\label{sec: related works}
\subsection{MTQE Models}
MTQE models estimate the quality of machine translated text \citep{rei2020comet}. There are two kinds: reference-free and reference-based \citep{comet22, comet23}. 

\paragraph{Reference-free} models are given a source text and a translated text as input. Their output is a scalar score between 0 to 1 that indicates the semantic equivalence between the source and translation (the closer it is to 1, the stronger the semantic equivalence) \citep{zhao2024handcraftedfeaturesllmsbrief}. Thus, during training, sentences that are semantically closer will output higher scores and those that are semantically less equivalent will receive lower scores.

\paragraph{Reference-based} models, on the other hand, are given a source text, translated text, and reference text. Their output is also a scalar between 0 to 1, and also indicate the semantic equivalence between source, translation, and reference. They rely on gold standard, human-annotated references off of which to base the numerical MTQE scores. During training, the model learns to output direct assessment scores to the given source, translated, and reference text \citep{comet22}.

We posit that due to MTQE models being trained on parallelly translated and/or human annotated data, they autonomously learn how to model non-compositional language. We can use such models as a form of weak distillation to teach models to translate non-compositional language effectively.

\subsection{Idiom Translation}
Idiom and proverb detection and generation have been studied extensively in literature \citep{cheng2024no, lai2024survey}. Some works focus on detection and generation \citep{wu2024refining, zhou2023non, he2024enhancingidiomaticrepresentationmultiple, conversational_idiom_detection}, while others work on creating high-quality curated datasets \citep{zeng2023iekgcommonsenseknowledgegraph, rezaeimanesh2024persianidioms, id10m, magpie}.



While detection and generation abilities are still improving, translation remains a problem \citep{cheng2024no}. Sources show that while proprietary, closed-source language models are able to translate idioms well, open-source language models have not yet reached SOTA results \citep{obeidat2024analyzing}. \citet{idiom_memorization} suggests language models do indeed know the semantic meaning behind idioms, they just need to be extracted properly. Previous work has mostly used prompting methods \citep{idiom_generation_prompting, rafatbakhsh2021development}. Rather than default to translating literal meanings \citep{rezaeimanesh2024persianidioms}, language models must be fine-tuned for translating semantic meanings behind idioms. Recent literature suggests that reinforcement learning can sharpen output distributions towards a specific task \citep{incentivising_reasoning}. \textit{We borrow this finding and use GRPO to improve a language model's ability to translate non-compositional language}.

\section{Improving Non-Compositional Translation}
\label{sec: methodology}

\subsection{Dataset Creation}
\label{sec: dataset}
Our evaluation spans two languages: Chinese and Hindi. We release our datasets, along with all of our code \href{https://github.com/agarwalishika/TranslatingIdioms}{here}. The same training and testing splits are used for all baselines, which allows for direct comparison between all methods.

\paragraph{Chinese.} We use the PETCI dataset \citep{tang2022petci} for Chinese–English idiom translation. We first run a preprocessing script on the original release that trims stray whitespace and removes rows where either the Chinese idiom or the English translation is empty, contains only whitespace, or is marked as missing. After cleaning 4,310 sentences in the original dataset, we obtain 1,623 valid idiom–translation pairs. We use 1,000 for training and 623 for testing.

\paragraph{Hindi.} We compile a dataset from multiple sources, combining mined Hindi–English idioms from the OPUS OpenSubtitles corpus \citep{opus} with additional high-quality synthetic pairs generated using GPT-5. We filter this collection by removing duplicates, discarding entries with incomplete or overly literal translations, and manually validating idioms that appear ambiguous or context-dependent. After cleaning, we select 1,000 valid Hindi–English idiom pairs, which we split deterministically into 800 training examples and 200 test examples. 

\begin{figure*}[t]
    \centering
    (a) \fbox{\includegraphics[width=0.18\linewidth]{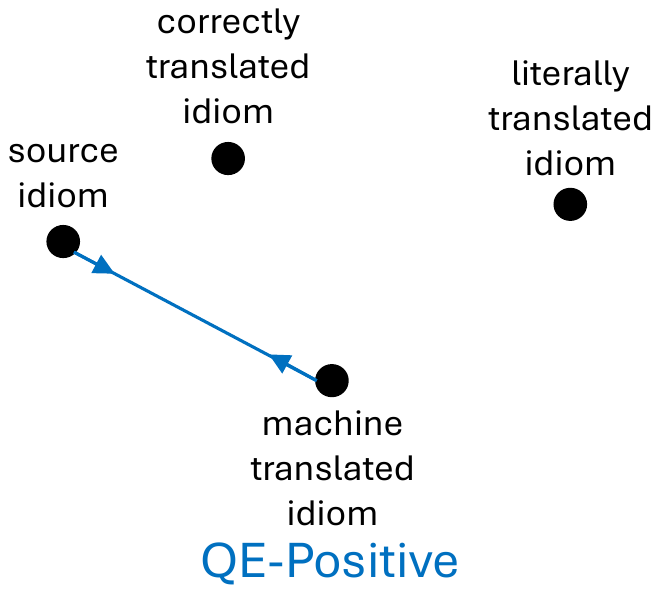}}
    (b) \fbox{\includegraphics[width=0.18\linewidth]{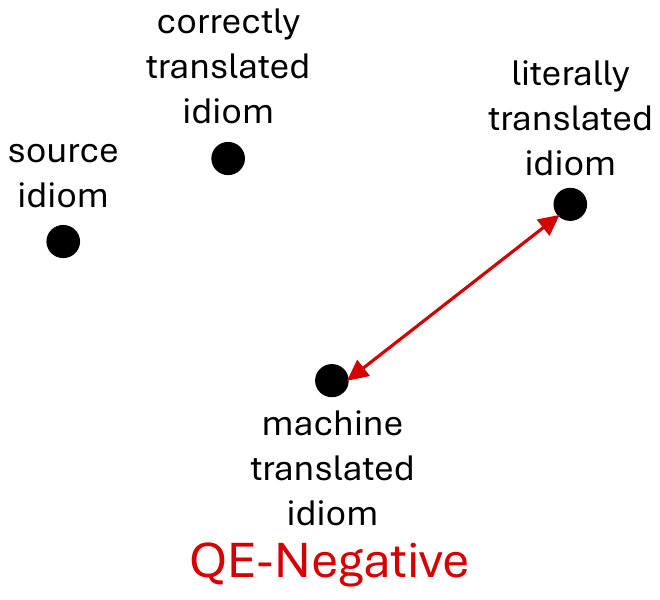}}
    (c) \fbox{\includegraphics[width=0.18\linewidth]{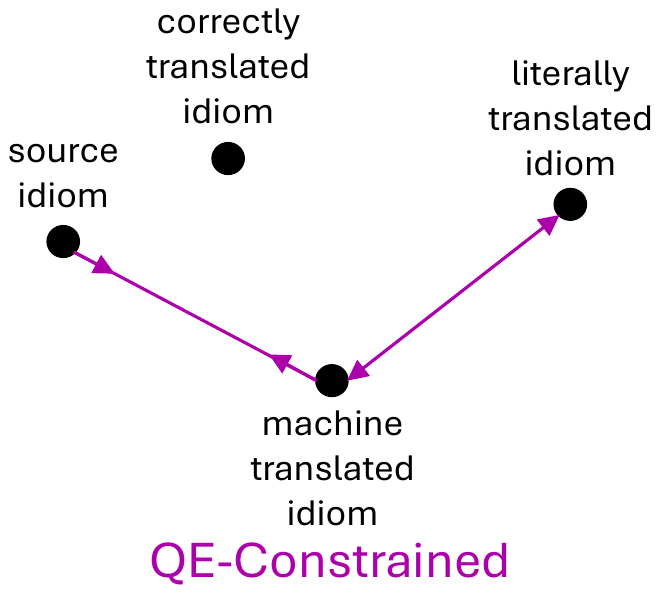}}
    (d) \fbox{\includegraphics[width=0.18\linewidth]{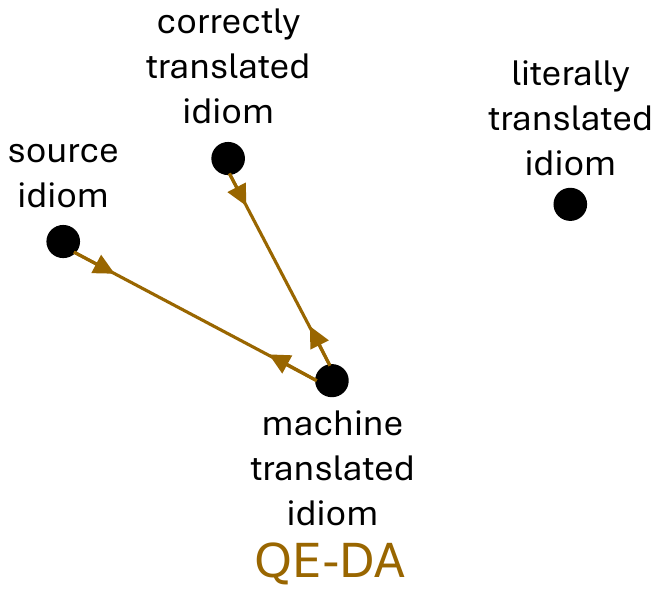}}
    \caption{Illustration of the distinction between all three GRPO-QE-* methods. (a) \textcolor{customblue}{QE-Positive} pulls semantically equivalent texts closer. (b) \textcolor{customred}{QE-Negative} pulls semantically inequivalent texts apart. (c) \textcolor{custompurple}{QE-Constrained} balances both. (d) \textcolor{custombrown}{QE-DA} uses a ground-truth reference translation to inform MTQE.}
    \label{fig: qe}
\end{figure*}

\subsection{Training-Based}
\label{sec: training-based}
We explore using GRPO \citep{grpo} for fine-tuning models to improve an LLM's translation abilities. In particular, the reward model is an MTQE model, and the LLM is rewarded based on how effective translations are based on the ground truth. As mentioned before, there are two kinds of MTQE models: reference-free and reference-based. Within these variations, we contrive different settings for rewards (see Figure \ref{fig: qe}):

\begin{enumerate}
    \item \textcolor{customblue}{\textbf{QE-Positive.}} QE models take in as input the source text (\texttt{src}) and the machine translated text (\texttt{mt}). In this setting, \texttt{src} is the input idiom in a different language, and \texttt{mt} is the LLM generated translation during training. This reward encourages LLMs to generate translations that are semantically equivalent to the original input idiom. We denote this as $\text{QE}_{\text{pos}}(\texttt{idiom}, \texttt{mt})$.
    
    \item \textcolor{customred}{\textbf{QE-Negative.}} Here, \texttt{src} is the literal meaning of the idiom (in English) and \texttt{mt} is, again, the generated translation during training. Note: both \texttt{src} and \texttt{mt} are in English in this setting. This is a creative misuse of MTQE. Although it is expected to have crosslingual outputs, it is ultimately a semantic similarity metric. In this case, we apply a negative reward for the MTQE between \texttt{src} and \texttt{mt}, encouraging the LLM to generate translations that are not semantically equivalent to the literal translations. We denote this as $\text{QE}_{\text{neg}}(\texttt{literal}, \texttt{mt})$.
    
    \item \textcolor{custompurple}{\textbf{QE-Constrained.}} The issue in the QE-Positive setting is that bad translations do not get \textit{discouraged}. The issue in the QE-Negative setting is that some idioms can be correctly literally translated in another language (for example: "plenty of fish in the sea" in English and "Hay más peces en el mar" in Spanish are correct, literal translations of each other). To bridge these gaps, we test a third setting. We simply assign the joint reward: $\text{QE}_{\text{pos}}(\texttt{idiom}, \texttt{mt}) - \text{QE}_{\text{neg}}(\texttt{literal}, \texttt{mt})$. This helps encourage the LLM to semantically translate the idiom, and discourage the LLM to literally translate.

    \item \textcolor{custombrown}{\textbf{QE-DA}}. This setting follows the \textcolor{customblue}{QE-Positive} setup, where \texttt{src} is the source idiom, and \texttt{mt} is the machine translated idiom, but it also receives a \texttt{ref}, which is the ground truth translated idiom. This reward will encourage models to translate idioms towards a specific target translation.
\end{enumerate}

\subsection{Training-Free Structured Prompting}
\label{sec: training-free}
We also develop a training-free idiom translation method designed to reduce literal-translation bias with just prompt engineering. The core idea is to dissect idiom translation into three explicit reasoning stages:

\begin{enumerate}
    \item \textbf{Idiomatic Explanation.}
    The model explains the idiom's figurative meaning in English, emphasizing cultural meaning over surface semantics, denoted by $E$ (prompt in Figure \ref{fig: idiomatic_explanation_prompt}, Appendix \ref{app: prompts}).

    \item \textbf{Literal Semantics.}
    The model provides a word-by-word literal translation of the idiom, yielding $L$, which helps to disentangle literal and figurative interpretations (prompt in Figure \ref{fig: literal_semantics_prompt}, Appendix \ref{app: prompts}).

    \item \textbf{Natural Idiomatic Translation.}
    Given both $E$ and $L$, the model produces a single fluent English expression that captures the idiomatic sense (prompt in Figure \ref{fig: natural_idiomatic_translation_prompt}, Appendix \ref{app: prompts}).
\end{enumerate}

Unlike decoding-based reranking or reinforcement learning, this method introduces no additional optimization, instead relies entirely on prompt engineering and explicit reasoning decomposition. This structured prompting approach aims to mitigate the literalism observed in the raw model.

\begin{figure*}[h]
    \centering
    \includegraphics[width=0.9\linewidth]{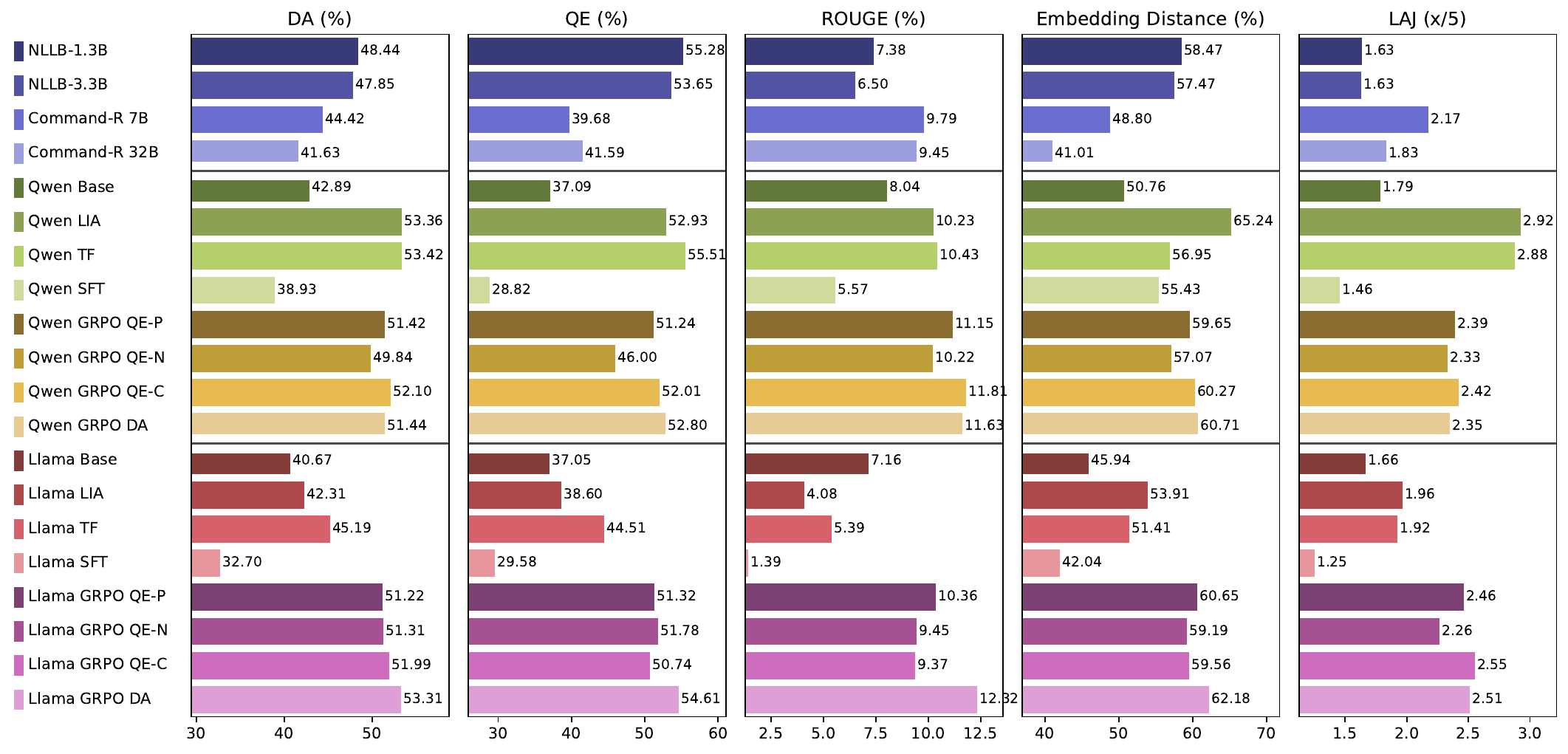}
    \caption{Evaluation of translation abilities of \textbf{Chinese idioms}. Here, we see that the LIA and \texttt{TrainingFree} (TF) baselines do well on Qwen-2.5-3B, but not on Llama-3.1-8B (hence is unreliable). The GRPO based methods are not only \textbf{performant}, but also \textbf{reliable}.}
    \label{fig: chinese_eval}
\end{figure*}

\begin{figure*}[h]
    \centering
    \includegraphics[width=0.9\linewidth]{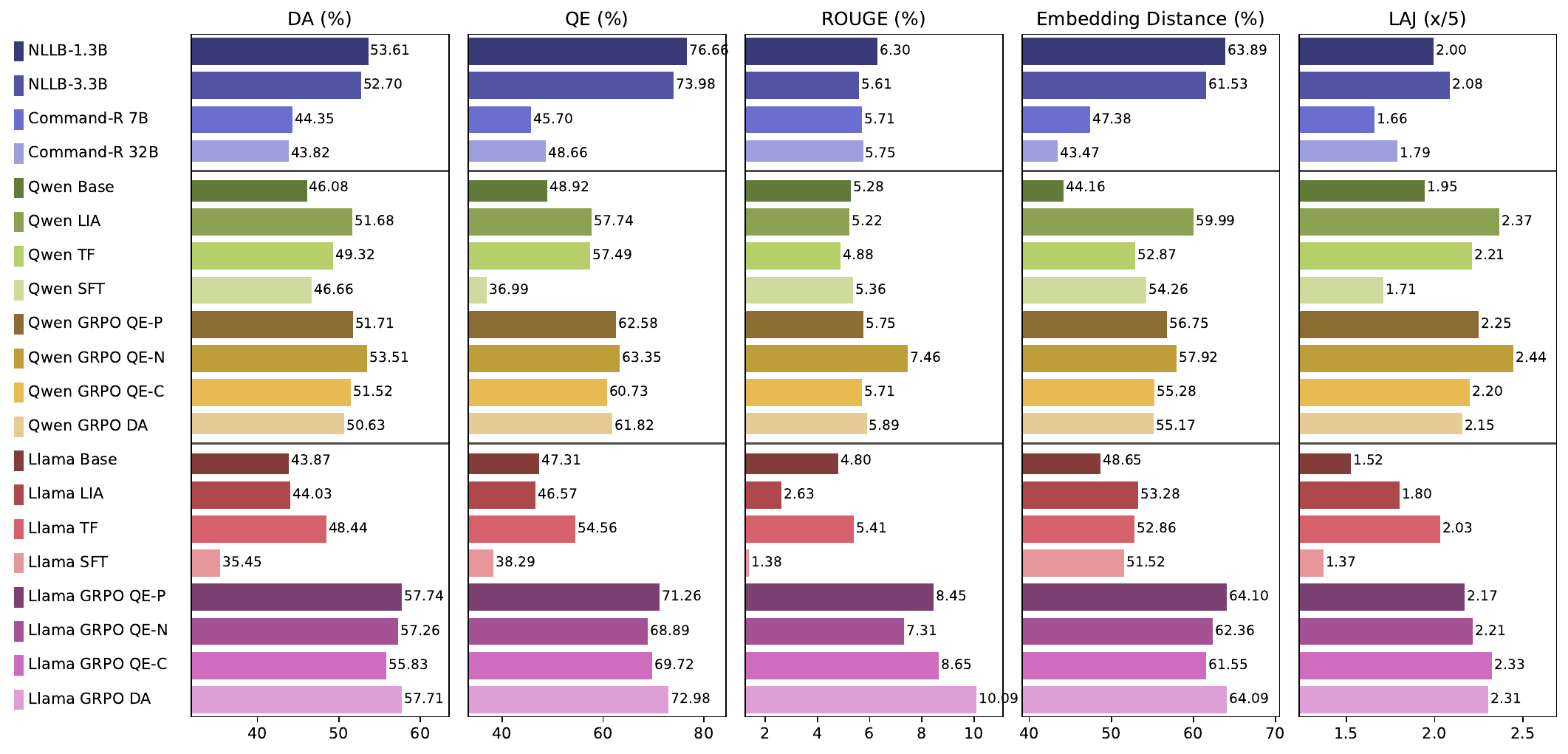}
    \caption{Evaluation of translation abilities of \textbf{Hindi idioms}. Here, we see that the core translation models (NLLB and Command-R) translate Hindi idioms well, but nChinese idioms (Fig. \ref{fig: chinese_eval}). The GRPO based methods are not only \textbf{performant}, but also \textbf{reliable}.}
    \label{fig: hindi_eval}
\end{figure*}

\begin{figure*}[h]
    \centering
    \includegraphics[width=0.9\linewidth]{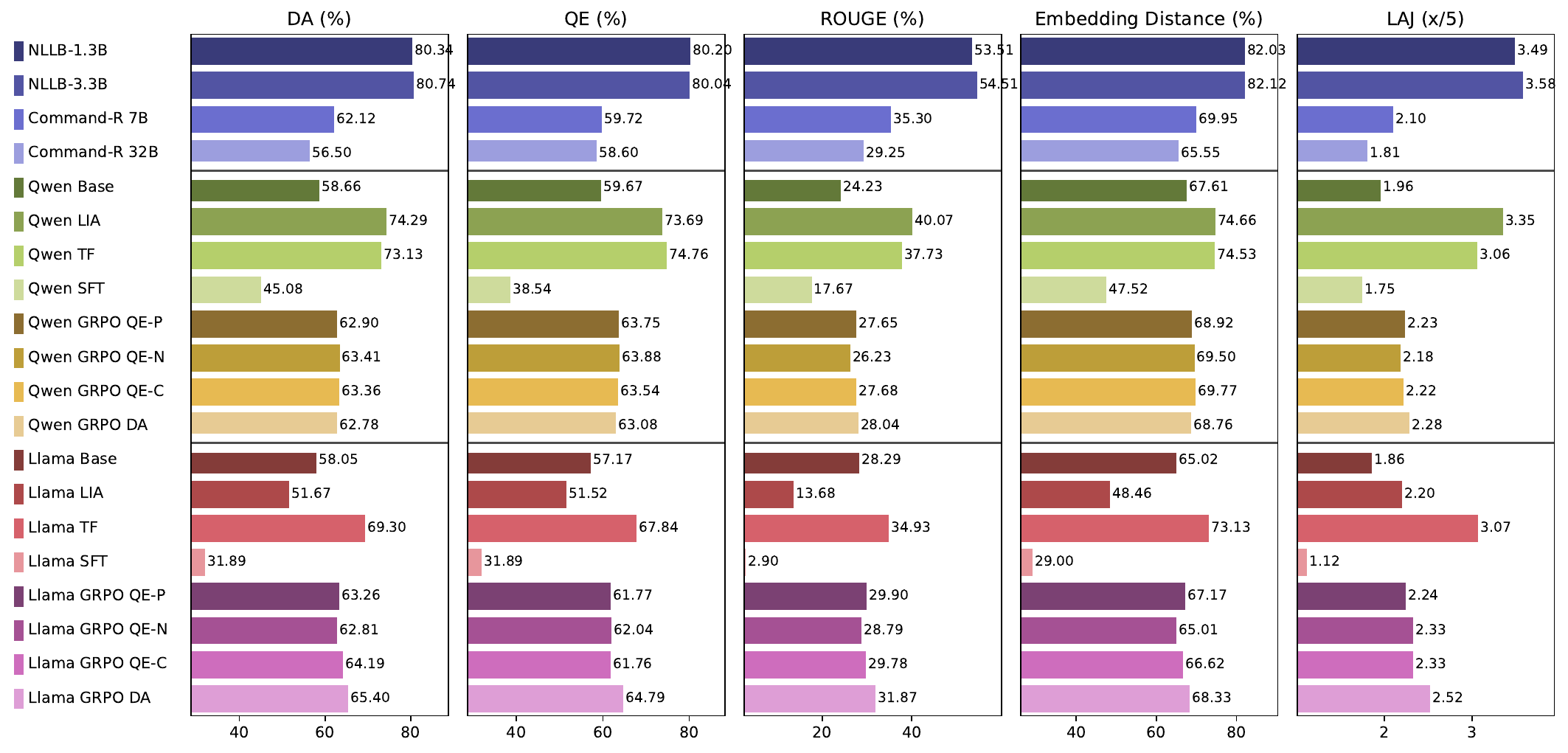}
    \caption{Evaluation of translation abilities of \textbf{regular Chinese sentences}. Here, it is shown that performance does not deteriorate when models are trained on idiomatic data.}
    \label{fig: opus_chinese_eval}
\end{figure*}

\begin{figure*}[h]
    \centering
    \includegraphics[width=0.9\linewidth]{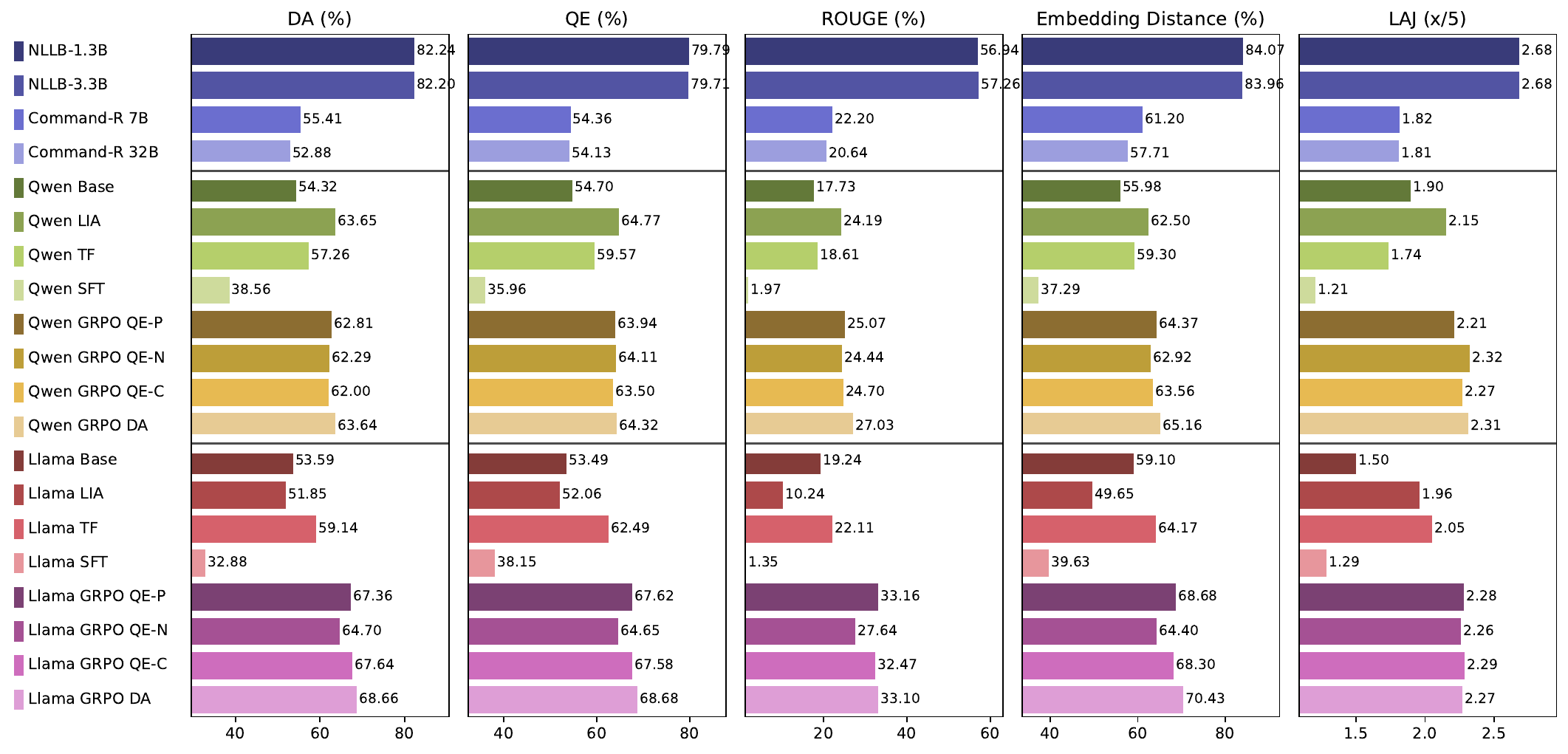}
    \caption{Evaluation of translation abilities of \textbf{regular Hindi sentences}. Here, it shows that performance does not deteriorate when models are trained on idiomatic data.}
    \label{fig: opus_hindi_eval}
\end{figure*}

\begin{figure*}[h]
    \centering
    \includegraphics[width=0.9\linewidth]{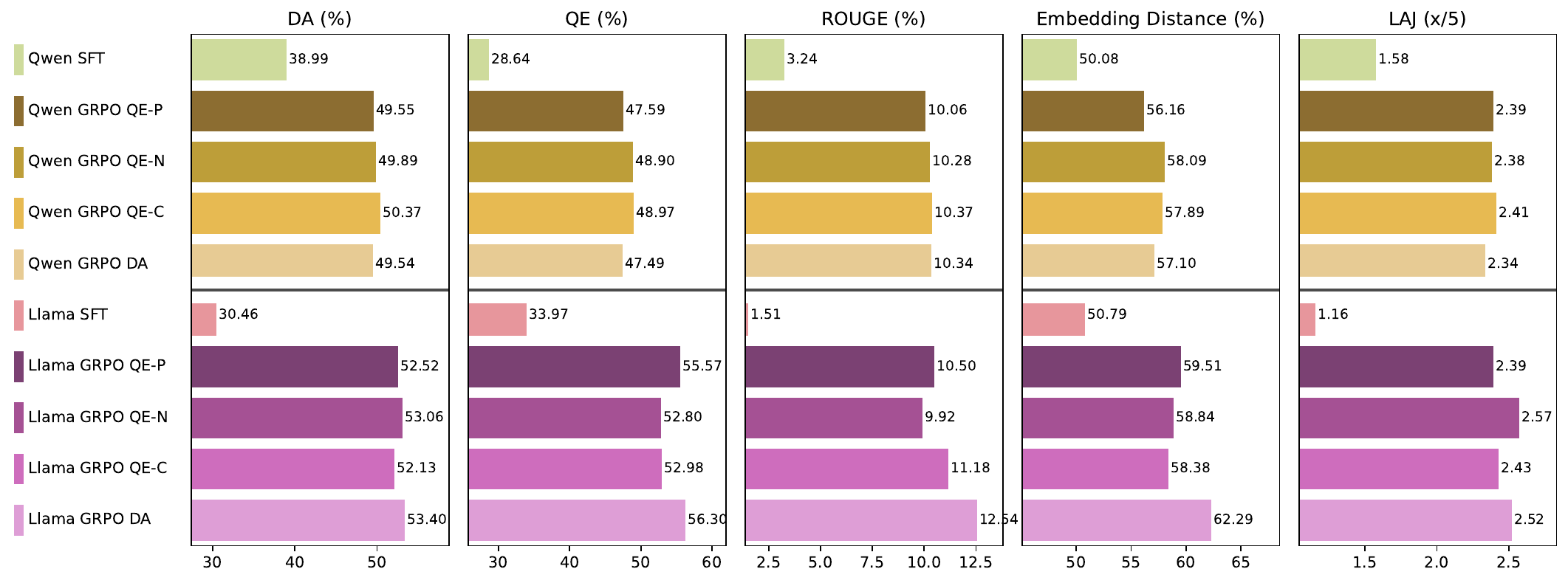}
    \caption{Evaluation of \textbf{Hindi-trained models translating Chinese idioms}. Per Fig. \ref{fig: chinese_eval}, Qwen-2.5-3B achieved DA: 42.89, QE: 37.09, ROUGE: 8.04, ED: 50.76, LAJ: 1.79; Llama-3.1-8B achieved DA: 40.67, QE: 37.05, ROUGE: 7.16, ED: 45.94, LAJ: 1.66. GRPO models outperform base models.}
    \label{fig: transfer_eval_hi2zh}
\end{figure*}

\begin{figure*}[h]
    \centering
    \includegraphics[width=0.9\linewidth]{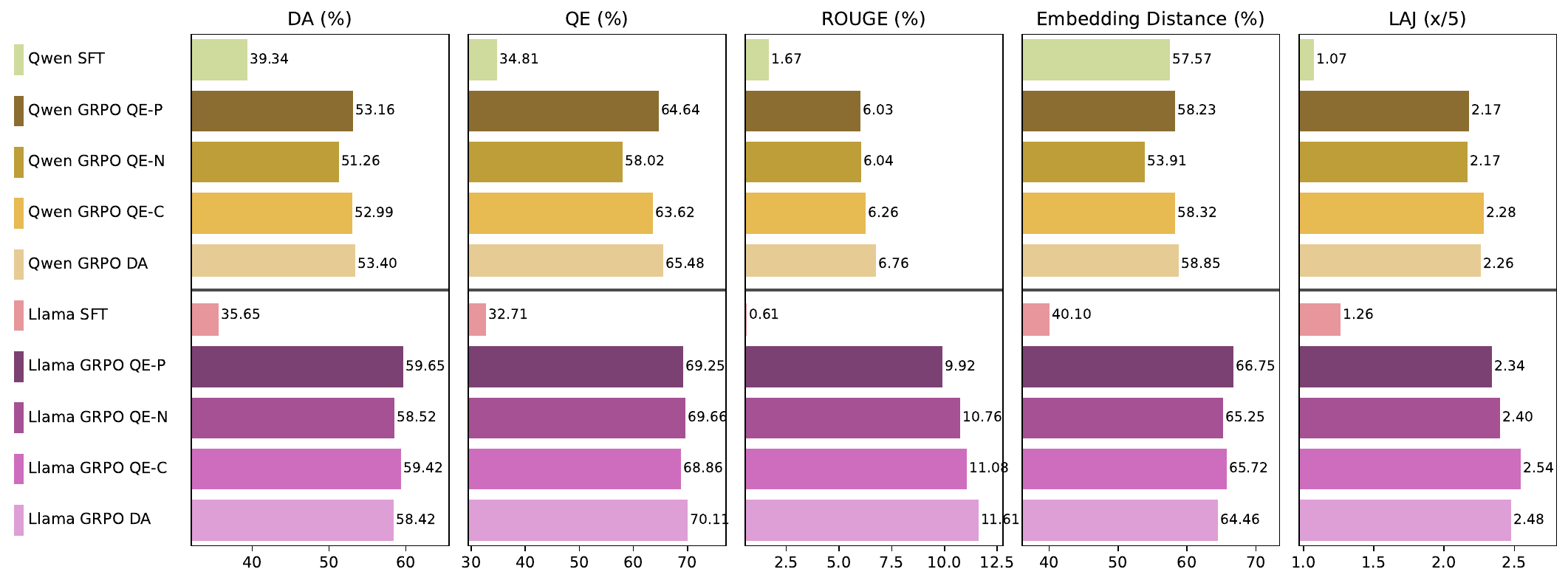}
    \caption{Evaluation of \textbf{Chinese-trained models translating Hindi idioms}. Per Fig. \ref{fig: hindi_eval}, Qwen-2.5-3B achieved DA: 46.08, QE: 48.92, ROUGE: 5.28, ED: 44.16, LAJ: 1.95; Llama-3.1-8B achieved DA: 43.87, QE: 47.31, ROUGE: 4.80, ED: 48.65, LAJ: 1.52. GRPO methods outperform base models.}
    \label{fig: transfer_eval_zh2hi}
\end{figure*}

\section{Experiments}
\label{sec: experiments}

\subsection{Setup}
\paragraph{Models.} To show the efficacy of our method, we use two small language models: (1) \texttt{Qwen/Qwen2.5-3B} (shorthand: Qwen) \citep{qwen3}, and (2) \texttt{meta-llama/Llama-3.1-8B} (shorthand: Llama) \citep{llama3}. We employ small LMs to showcase the efficacy of our method compared to larger LMs. For all model inference, we use a temperature of 0.3. For SFT, we train models for 3 epochs. For GRPO, we use the \texttt{verl} library \citep{verl}, with 4 completions per group, and 5 epochs. All other settings can be found in our \href{https://github.com/agarwalishika/TranslatingIdioms}{Github repository}.

\paragraph{Metrics.} Our evaluation contains a variety of metrics to test different aspects of the translation quality. First off, we report the scores of the same MTQE models that we use as reward models for GRPO, to show the final reward of the tuned models. \textbf{DA} are the Direct Assessment scores (specifically, the \texttt{Unbabel/wmt22-comet-da} model), and \textbf{QE} are the Quality Estimation scores (specifically, the \texttt{Unbabel/wmt22-cometkiwi-da} model). We evaluate the n-gram similarity between the predicted translation and the ground truth translation with \textbf{ROUGE}. However, ROUGE is too rigid a metric that measures the n-gram overlap ration and does not measure for semantic similarity --  hence, we also use embedding models to embed the semantic similarity of the predicted translation and ground truth translation; we, then, report the cosine distance between these embeddings (i.e., \textbf{Embedding Distance}). Finally, to make our evaluation comprehensive, we employ an LLM-as-a-Judge (abbreviated to \textbf{LAJ}) to score the semantic similarity between the predicted and ground truth translation. The particular LLM-as-a-Judge we use is the Prometheus-7v-V2.0 model \citep{kim2024prometheus}.

\paragraph{Baselines} We employ a variety of baseline methods. First, we use core translation models: sequence-to-sequence (\textbf{NLLB} \citep{nllb}) and autoregressive (\textbf{Command-R} \citep{command_r}). To compare our training-free method, we use the \textbf{LIA} (Language Model Based Idiom Alignment) \citep{lia}\footnote{The same paper also proposes SIA (Semantic Idiom Alignment), but their setting requires an external database of idioms to retrieve from, which is then fed into SIA to select highly matching translations. We do not choose to compare against this particular baseline as we do not assume access to an external database.}. To compare our training-based method, we fine-tune models using supervised fine-tuning (\textbf{SFT}) using a cross-entropy loss.

\paragraph{Non-Idiomatic Performance} In order to ensure our idiom-specific methods (especially, the training-based) do not deteriorate in translating non-idiomatic text, we measure the performance of our methods on non-idiomatic text translation. We use the same metrics as mentioned above, and we employ the Opus-100 \citep{opus} dataset. This dataset contains pairs of source and machine translated text pairs. We use the \texttt{en-zh} and \texttt{en-hi} pairs from the dataset, and use 400 randomly selected data points from the test splits of the Opus dataset for our evaluation.

\subsection{Results and Analysis}
\paragraph{Idiom Translation.} Figures \ref{fig: chinese_eval} and \ref{fig: hindi_eval} contain the evaluation results on Chinese and Hindi idioms, respectively. The NLLB and Command-R models tend to be better at the Qwen and Llama base models (they score an average of 2.91 (Qwen) and 5.03 (Llama) points\footnote{To aggregate these metrics into one reportable score, we take the average of all 5 metrics: DA, QE, ROUGE, Embedding Distance and LAJ. The first four are on the same scale, while LAJ is on a scale of 1-5. To calibrate them all on the same scale, we multiply the LAJ score by 20. Thus, to calculate the performance of a particular method, we use the following formula: $p = (\text{DA} + \text{QE} + \text{ROUGE} + \text{EmbeddingDistance} + 20*\text{LAJ}) / 5$. To find the difference in performance between baseline $A$ and method $B$, we calculate $p_A - p_B$.} higher for Chinese translation, and 4.80 (Qwen) and 6.48 (Llama) points higher for Hindi translation. Still, the results show that \textit{training-free/-based methods can improve flexible language models so they can perform on-par with translation models} that have dedicated machine translation or parallelly translated datasets\footnote{According to the model cards \citep{qwen3, llama3}, the training sets have multilingual data, but the data is not necessarily parallelly translated in other language. These multilingual data are generally for multilingual question-answering, reasoning, and knowledge tasks.}.

Comparing the training-free methods (LIA and \texttt{TrainingFree} (denoted by TF in the figures)), we see that Qwen performs better with LIA (LIA is 1.65 points better than \texttt{TrainingFree}), while Llama performs better with \texttt{TrainingFree} (it is 2.18 points better than LIA). These inconsistent results show that prompting is not a reliable method, motivating the need for fine-tuning.

Supervised fine-tuning the models consistently cause models to perform worse than even the base models -- this is because the models are being trained for outputting particular sentences, rather than understanding semantic meaning of sentences. Thus, \textit{using RL is our final approach, which significantly outperforms the base and SFT models}: The \textcolor{customblue}{QE-Positive} models outperform the base and SFT models by 13.23 and 18.80 points, respectively; the \textcolor{customred}{QE-Negative} models outperform them by 13.20 and 18.77 points, respectively; the \textcolor{custompurple}{QE-Constrained} models outperform them by 13.64 and 19.22 points, respectively; finally, the \textcolor{custombrown}{QE-DA} models outperform them by 14.60 and 20.18 points, respectively. \textbf{Overall, the RL models bring an absolute point improvement of 13.67 points in idiom-translation ability.}

\paragraph{Performance on non-idiomatic sentence translation.} Figures \ref{fig: opus_chinese_eval} and \ref{fig: opus_hindi_eval} show the results of translation abilities of non-idiomatic text translation in Chinese and Hindi respectively. These tables show that \textbf{the performance on general, mostly non-idiomatic text translation not only maintains, but also improves with the GRPO models, compared to base (average of 8.39 points) and SFT (average of 25.07 points)}. Of course, the translation models are better performing (the best translation model is 7.90 points better than the best GRPO model). We hypothesize that the GRPO-tuned language models have improved semantic representation across languages, so they could ultimately outperform the translation models on reasoning, knowledge, and question-answering tasks.

\paragraph{Effects of language-specific training.} Figures \ref{fig: transfer_eval_hi2zh} and \ref{fig: transfer_eval_zh2hi} contain the evaluation results for the transfer settings. Comparing the results from Figures \ref{fig: hindi_eval} and \ref{fig: transfer_eval_zh2hi} versus Figures \ref{fig: chinese_eval} and \ref{fig: transfer_eval_hi2zh}, there do exist significant transferability capabilities between the trained models. In particular, models trained on Chinese idioms but evaluated on Hindi idioms perform 8.04(Qwen)/15.73(Llama) points \textit{better} than base models and 0.38(Qwen)/1.81(Llama) points \textit{better} than models trained on Hindi idioms. On the contrary, models trained on Hindi idioms but evaluated on Chinese idioms perform 7.73(Qwen)/12.70(Llama) points \textit{better} than base models, Qwen performs -1.31 \textit{worse} than models trained on Chinese, and Llama performs 0.76 points \textit{better}. The overall boost of 8.62 absolute points in translation improvement shows that \textbf{training on one language does not hinder the performance on other languages}.

\paragraph{Effect of reward model.} Even though we had broken down the rewards into four, \textbf{there are no noticeable effects of each reward}. This is encouraging because they all have different kinds of supervision. \textcolor{customblue}{QE-Positive} requires just an input idiom, which can be extracted from large corpora with idiom detectors. \textcolor{customred}{QE-Negative} and \textcolor{custompurple}{QE-Constrained} requires a source \textit{and} a literal translation, which might be a bit expensive to get. Although, costs can be avoided by using a small enough language model that reliably generates literal translations. \textcolor{custombrown}{QE-DA}, however, is the most expensive as it requires a ground truth translation of idioms -- obtaining this is expensive, as it necessitates annotation from either humans or closed-source large language models. \textbf{Since the least expensive reward \textcolor{customblue}{QE-Positive} does not lag far behind the most accurate reward \textcolor{custombrown}{QE-DA}, we are able to showcase the robustness of using MTQE rewards.}

\section{Conclusion}
In this work, we explore structured approaches to improving non-compositional language translation. As our main contribution, we present GRPO-based fine-tuning using MTQE models as reward models. Our experimentation uncovers three results: (1) idiom translation abilities increase by an average of 13.67 absolute points over base models across languages and architectures, (2) non-idiomatic translation abilities are implicitly improved by 8.39 absolute points, and (3) cross-lingual translation abilities are also improved by 5.73 absolute points. These results show that MTQE rewards effectively distill their multilingual language embedding capabilities to language models: they improve semantic relationships in multilingual language models, are reliable to not compromise current language model abilities, and enables models to generalize to other languages meaningfully. These results encourage future work to understanding the complementary nature of mapping semantics in various languages for improving multilingual language modeling.

\section{Limitations}
While MTQE models can handle a broad variety of languages and have shown to be aligned with human preferences, our method is upper-bounded by the performance of MTQE models -- the GRPO-trained translation models can only be as good as the MTQE models are. Plus, MTQE models also require many  parallel data to be trained. Furthermore, RL is an expensive algorithm in general -- even after choosing small model sizes and small datasets, training took around 6-12 hours to train on 4 NVIDIA H100s, which is not accessible. Future work involves understanding how to reduce the computational cost to train better multilingual models.

\bibliography{acl_lualatex}

\appendix

\section{Prompts}
\label{app: prompts}
Tables \ref{fig: idiomatic_explanation_prompt}, \ref{fig: literal_semantics_prompt}, and \ref{fig: natural_idiomatic_translation_prompt} contain the prompts for the \texttt{TrainingFree} method described in Section \ref{sec: methodology}.2.

\begin{figure*}[h]
  \centering
  \begin{tcolorbox}[
    colback=white!5!white,
    colframe=black!75!black,
    title=Idiomatic Explanation Prompt,
    boxrule=0.3mm,
    width=\textwidth,
    arc=1.5mm,
    auto outer arc
  ]

Explain the meaning of the following \{lang\} idiom in English. \\
- Audience: educated readers; be concise (<= 2 sentences). \\
- Do not translate word-by-word; provide the **idiomatic sense**. \\

Idiom: \{idiom\}
  \end{tcolorbox}
  \caption{Prompt to elicit Idiomatic Explanation, which is Step 1 in our training-free, structured prompting approach outlined in section \ref{sec: training-free}.}
  \label{fig: idiomatic_explanation_prompt}
\end{figure*}

\begin{figure*}[h]
  \centering
  \begin{tcolorbox}[
    colback=white!5!white,
    colframe=black!75!black,
    title=Literal Semantics Prompt,
    boxrule=0.3mm,
    width=\textwidth,
    arc=1.5mm,
    auto outer arc
  ]

Provide a **literal, word-by-word** English translation for the following \{lang\} idiom. \\
- Keep it terse and faithful to each component. \\
- No commentary, just the literal gloss. \\

Idiom: \{idiom\}
  \end{tcolorbox}
  \caption{Prompt to elicit Literal Semantics, which is Step 2 in our training-free, structured prompting approach outlined in section \ref{sec: training-free}.}
  \label{fig: literal_semantics_prompt}
\end{figure*}

\begin{figure*}[h]
  \centering
  \begin{tcolorbox}[
    colback=white!5!white,
    colframe=black!75!black,
    title=Natural Idiomatic Translation Prompt,
    boxrule=0.3mm,
    width=\textwidth,
    arc=1.5mm,
    auto outer arc
  ]

Produce a **natural English idiomatic translation** given:\\
(1) An idiom explanation (idiomatic meaning) and \\
(2) A literal word-by-word gloss. \\

Rules: \\
- Output a single short English phrase/sentence that a native speaker would actually say. \\
- Prefer clarity and naturalness over literalness. \\
- No extra commentary.\\
\\
Idiom: \{idiom\} \\
Explanation: \{explanation\} \\
Literal: \{literal\} \\
Result: 
  \end{tcolorbox}
  \caption{Prompt to elicit Natural Idiomatic Translation, which is Step 2 in our training-free, structured prompting approach outlined in section \ref{sec: training-free}.}
  \label{fig: natural_idiomatic_translation_prompt}
\end{figure*}

\section{Licenses}
All datasets are publicly available and do not contain any personally identifiable information. The PETCI dataset is under the Creative Commons 4.0 license, the Open Subtitles dataset is under the GNU General Public License, the NLLB and Command-R models are under the Creative Commons Attribution Non Commercial 4.0 license, the Qwen model is under the Apache 2.0 license, and the Llama model is under the Llama license.

\section{Human Annotators}
In creating the dataset, we used human annotators to verify the validity of the dataset. The authors of this paper were appropriate enough to create this dataset as they knew either Chinese or Hindi, and they were unbiased in their dataset verification process. We wrote up guidelines to make sure the idiomatic data was not "too literal" or "overly ambiguous" of whether it had a figurative or literal meaning.

\section{Usage of AI Assistants}
There were only two things AI Assistants were used for: (1) writing code for plotting figures (after processing the data ourselves, we described to ChatGPT the format of the figures that we wanted and asked it how to add a color scheme), and (2) revising small chunks of text like the title, the abstract, and caption figures (we had already written the title and abstract, but we instructed Claude to refine the writing by making it clearer and shorter). No AI Assistants were used for any of the other tasks: paper writing, code writing, results analysis, or ideation (or even, this section).

\end{document}